# Pathological Pulmonary Lobe Segmentation from CT Images using Progressive Holistically Nested Neural Networks and Random Walker

Kevin George, Adam P. Harrison, Dakai Jin, Ziyue Xu⋆, and Daniel J. Mollura

National Institutes of Health, Bethesda, MD, USA

**Abstract.** Automatic pathological pulmonary lobe segmentation(PPLS) enables regional analyses of lung disease, a clinically important capability. Due to often incomplete lobe boundaries, PPLS is difficult even for experts, and most prior art requires inference from contextual information. To address this, we propose a novel PPLS method that couples deep learning with the random walker (RW) algorithm. We first employ the recent progressive holistically-nested network (P-HNN) model to identify potential lobar boundaries, then generate final segmentations using a RW that is seeded and weighted by the P-HNN output. We are the first to apply deep learning to PPLS. The advantages are independence from prior airway/vessel segmentations, increased robustness in diseased lungs, and methodological simplicity that does not sacrifice accuracy. Our method posts a high mean Jaccard score of $0.888 \pm 0.164$ on a held-out set of 154 CT scans from lung-disease patients, while also significantly ($p < 0.001$) outperforming a state-of-the-art method.

## 1 Introduction

Due to the prevalence of lung disease worldwide, automated lung analysis tools are a major aim within medical imaging. Human lungs divide into five relatively functionally independent lobes, each with separate bronchial and vascular systems. This causes several lung diseases to preferentially affect specific lobes [1]. As a result, pathological pulmonary lobe segmentation (PPLS) is an important step toward computerized disease analysis tools.

There are several anatomical challenges to PPLS. In healthy lungs, lobar borders are typically defined by visible fissures. However, fissures are often incomplete, *i.e.*, they fail to extend across the full lobe boundary [2]. Thus, fissure detection and lobar boundary detection are not always equivalent problems. Even when fissures are visible, they can vary in thickness, location, and shape due to respiratory and cardiac motion [3]. Furthermore, accessory fissures, structures visually identical to major fissures but that do not actually divide lobe regions, are common [4]. Finally, although PPLS provides the greatest clinical utility in pathological lungs, lung diseases often completely obscure fissures.

⋆ Corresponding author: ziyue.xu@nih.gov. This work is supported by the Intramural Research Program of the National Institutes of Health, Clinical Center and the National Institute of Allergy and Infectious Diseases. Support also received by the Natural Sciences and Engineering Research Council of Canada. We also thank Nvidia for the donation of a Tesla K40 GPU.

For these reasons, most of prior art relies on pulmonary vessel/airway segmentations or on semi-automated approaches [1]. Bragman *et al.*'s work is an exemplar of the former category [5], achieving good accuracy, but with a relatively complex scheme involving pulmonary vessel and airway segmentations, in addition to population-based models of fissure location and surface fitting. Apart from the complexity involved, pulmonary airway and vessel segmentation are not always reliable, especially in the presence of pathologies. This is why Doel *et al.* [6], for example, attempt to minimize dependence on the quality of these priors by using them mainly to initialize fissure detection in their method.

Only two lobe segmentation methods have been proposed that are both automatic and non-reliant on prior airway/vessel segmentations [7, 8]. However, both methods have noticeable limitations. Ross *et al.* [7] use a manually defined atlas, which is laborious to create, prone to poor results when tested on highly variable pathological lungs, and increases execution time [1]. Pu *et al.* [8] also achieve lobe segmentation without reliance on prior airway/vessel segmentations, but they report less success in cases with incomplete fissures, large accessory fissures, and where lobe boundaries have abnormal orientations. Finally their method is only validated on healthy volumes and only qualitative results were reported. So far, no deep learning based algorithm has been proposed.

To fill this gap, we develop a PPLS approach that combines a state-of-art deep fully convolutional network (FCN) with the random walker (RW) algorithm. Observing that FCNs have proven highly powerful for segmentation, we use the recently proposed progressive holistically-nested network (P-HNN) architecture [9] to segment lobar boundaries within the lung. Importantly, P-HNN is able to perform well even when fissures are incomplete. The P-HNN probability map then is used to generate seeds and edge probabilities for the RW algorithm [10], which produces a final lobe segmentation mask. We call our method P-HNN+RW, which is the first to apply deep learning to PPLS. Importantly, P-HNN+RW only requires a lung mask to function and remains simple, fast, and efficient, yet robust. We test P-HNN+RW on the highly challenging Lung Tissue Research Consortium (LTRC) dataset [11], which includes CT scans of patients suffering from chronic obstructive pulmonary disease (COPD) and interstitial lung disease (ILD), and demonstrate improvement over a state-of-the-art method [6] despite being much simpler in execution. On a held-out test set of 154 volumes, the largest analysis of lobe segmentation to date, we achieve a high Jaccard score (JS) of $0.888 \pm 0.164$.

## 2  Method

Fig. 1 illustrates a flowchart of our PPLS pipeline. To avoid relying on prior vessel/airway segmentation, we instead employ a principled combination of the P-HNN deep-learning architecture [9] with the celebrated RW algorithm [10]. The only pre-requisite is a lung mask, where reliable methods do exist [9].

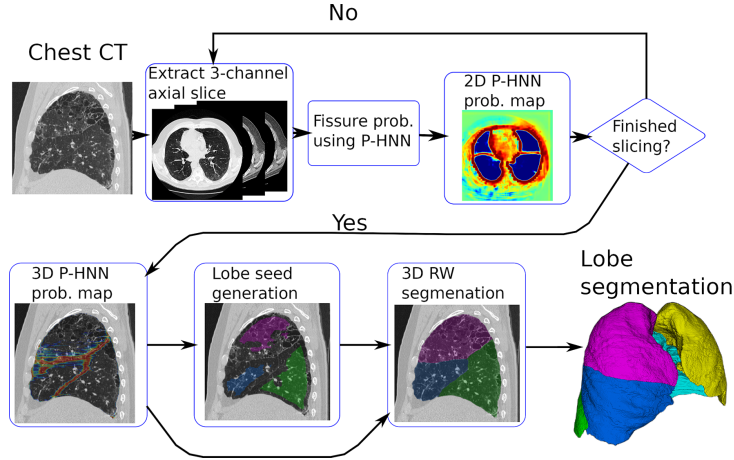

Fig. 1: Flowchart of P-HNN+RW's operation.

### 2.1 Lobar Boundary Segmentation

Given a volume and a lung mask, if a reliable boundary map between lobes were calculated, segmenting the five lobes would be relatively straightforward. Unfortunately, lobar boundaries are difficult to detect, given the common occurrence of incomplete features and other confounding factors, causing much of the prior art to rely on supplemental information given by airway or vessel segmentations. Reliably segmenting lobar boundaries directly from attenuation values remains attractive as it would avoid this considerable complexity. The recent prominence and success of FCNs in highly challenging segmentation tasks has opened up the possibility of more reliable lobar boundary segmentation.

Along those lines, our method's first step is to apply an FCN model to lobar boundary segmentation. Without loss of generality, we adopt the recent P-HNN model [9], which has been effectively applied to pathological lung segmentation. Fig. 2 illustrates the P-HNN model, which is an adaptation of the highly successful HNN model [12] that introduced the concept of deeply supervised side-outputs to FCNs. Importantly, the model avoids the complicated upsampling pathways used in recent FCN architectures, instead using progressive multipath connections to help address the well-known coarsening resolution problem of FCNs. As such, the model remains straightforward and relatively simple, making it attractive for PPLS.

Regardless of the specific FCN used, the model functions by training on a set of data $S = \{(X_n, Y_n), n = 1\ldots, N\}$, where $X_n = \{x_j^{(n)}, j = 1\ldots, |X_n|\}$ and $Y_n = \{y_j^{(n)}, j = 1\ldots, |X_n|\}$, $y_j^{(n)} \in \{0, 1\}$ represent the input and binary ground-truth images, respectively. The latter denotes which pixels are located on a boundary region, i.e., $x_j \in \Omega_{\text{bound}}$, where we do not discriminate based on which exact lobes are being separated. Given an image and a set of parameters,

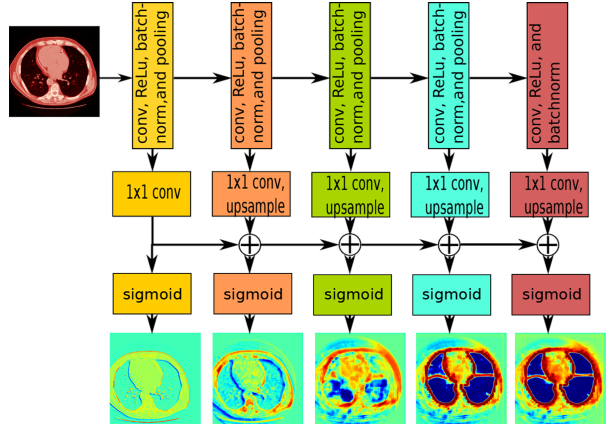

Fig. 2: Lobar boundary segmentation with P-HNN. Like HNN [12], the P-HNN model computes side-outputs, which are all connected to a loss function. Unlike HNN, P-HNN does not have a fused output, but constrains the side-outputs to progressively improve upon the previous output. Regions outside of the lung mask are ignored in training. Figure adapted with permission [9]

$\mathbf{W}$, the network then estimates

$$p_j = P\left(x_j \in \Omega_{\text{bound}} | X_n; \mathbf{W}\right).  \quad (1)$$

We perform training and inference on 2D axial slices, choosing the 2D domain for two main reasons. For one, current 3D FCN solutions require a sliding box, limiting the spatial context, which we posit is important for delineating lobar boundaries, especially when fissures are incomplete or obscured. In addition, sliding boxes add to the complexity and inference time, which we aim to minimize. Second, due to memory constraints, 3D FCNs are less wide and deep than their 2D counterparts. Given that excellent performance has been achieved in segmenting anatomical structures in 2D [9, 13, 14], we opt for a 2D solution that requires no compromise in network depth and width and remains simpler in application. Regions outside the lung mask are ignored in training.

### 2.2 3D Random Walker

Although the P-HNN model can provide a relatively reliable lobar boundary map, the output is still noisy, often resulting in gaps. Thus, the final segmentation cannot be obtained from the boundary maps alone, meaning a subsequent processing step, one that is robust to noise, is required. Conventionally, this can be done by simply fitting a surface to potential fissure locations [8], but this can be inefficient and error-prone. Instead, we combine all axial slices from the P-HNN output and execute 3D RW segmentation [10]. Typically used in interactive settings, the RW algorithm can be also used whenever there are clearly defined seeds, *i.e.*, regions constrained to fall in a specific segmentation mask.

We choose RW over other graph-based methods, *e.g.*, dense CRFs, due to its globally optimal solution, efficiency, and simplicity.

Formally, assume we have five sets of seed points, $R_k$, $k = 1 \ldots 5$, corresponding to regions that definitively belong to one of five lobes. If we partition the seed points into $S_f = R_1$ and $S_b = \{R_m, m = 2 \ldots 5\}$, then the pseudo-probability that the unseeded points belong to lobe 1 is calculated by minimizing the following energy functional:

$$E(\mathbf{y}) = \sum_{e_{ij}} w_{ij}(y_i - y_j)^2 \quad \text{s.t.,} \quad \begin{matrix} y_i = 1, \, x_i \in S_f \\ y_i = 0, \, x_i \in S_b \end{matrix}, \qquad (2)$$

where $y_j$ are the resulting probabilities and $w_{ij}$ denotes the weights penalizing differences across the edges, $e_{ij}$, connecting each pixel to its neighbors. We use a 6-connected neighborhood and ignore regions outside the lung mask. Minimizing (2) can be done linearly [10]. For all other lobes, the process is repeated, except that different lobes must be treated as foreground, *i.e.*, $S_f$, with the rest as background, *i.e.*, $S_b$. Because all pseudo-probabilities sum to 1 [10], the last lobe's segmentation can be calculated from the first four, saving one RW execution.

To minimize (2), seed regions and edge weights must be generated. To compute seed regions, we first invert the P-HNN boundary mask and then iteratively erode it until we obtain five fully connected regions, one for each lobe. Region centroids are used to identify specific lobes and a stopping threshold is used in cases where five regions cannot be determined. This iterative erosion approach breaks connections between lobes caused by incomplete or weak P-HNN boundaries, and is stopped at the point of the largest possible seeds for the RW. These enlarged seed regions reduce the size of the linear system, keeping a fully 3D RW implementation within memory constraints.

With the seeds generated, we calculate edge weights based on the P-HNN output:

$$w_{ij} = \exp(-\beta(p_i - p_j)^2)), \qquad (3)$$

where $p_{(.)}$ is obtained from (1) and $\beta$ is a global parameter analogous to a diffusion constant in physical systems, whose value is chosen based on validation subsets. Thus, the RW penalizes lobar mask leakage across the P-HNN boundaries, while producing a result robust to gaps or noise.

## 3  Experiments and Results

We train on axial slices from 540 annotated volumes from the LTRC dataset [11], testing on a held-out test set of 154 volumes. Importantly, every volume exhibits challenging attenuation patterns from ILD and/or COPD; thus, we test on pathological cases where PPLS could provide clinical utility. We rescale each slice to a 3-channel 8-bit image using three windows of $[-1000, 200]$, $[-160, 240]$, and $[-1000, -775]$ Hounsfield units. We fine-tune from the ImageNet VGG-16 model [15] and stop training after roughly 2 epochs. Parameters for the erosion step and RW were selected based on a validation set of 60 volumes. During inference, the full pipeline takes on average $4 - 8$ minutes for a typical volume using a Tesla K40 GPU and 12-core processor. To our knowledge, this represents the largest evaluation of a PPLS technique published to date.

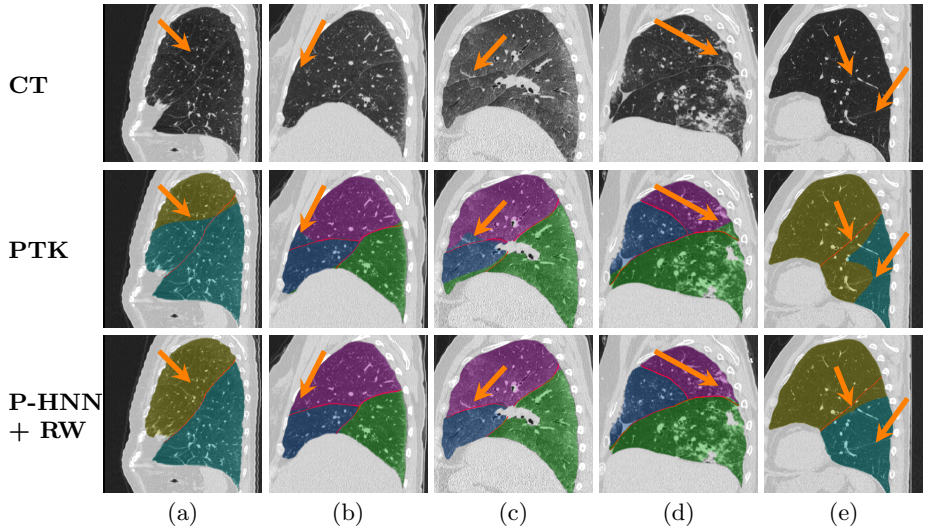

Fig. 3: Example PPLSs with ground-truth boundaries rendered as red lines. (a) P-HNN+RW correctly segments the left lung lobes, while PTK follows an erroneous boundary. (b) P-HNN+RW successfully handles an incomplete fissure. (c) PTK over-segments one lobe. (d) P-HNN+RW, unlike PTK, successfully differentiates fibrosis from a lobe boundary. (e) Despite no visual fissure, P-HNN+RW is able to use context to infer a more reasonable lobe boundary. Note, accessory fissure is ignored.

To measure P-HNN+RW's performance, without the lung mask as a confounding factor, we first test using the ground-truth lung mask. Tbl. 1 depicts the mean JSs and their standard deviation for each lobe over the full test set. P-HNN+RW is strongest in the left lung and weakest in the right middle lobe, a consistent trend among current methods [5, 16]. Importantly, we obtain a high overall Jaccard score of $0.888 \pm 0.164$, which compares very favorably with the highest performing published methods tested on similarly challenging pathological lungs [5, 16, 17], but without requiring vessel/airway segmentations, groupwise priors, and other complex processes.

We also directly compare against prior art. Since no other deep-learning method has been published, we instead compare against the pulmonary toolkit (PTK)'s [6] automatic and non-deep-learning scheme. For fair comparison, we regenerated our results using the PTK lung masks. The PTK failed to produce a final lobe segmentation for 40 volumes. For the remaining 114 volumes, Fig. 4 depicts cumulative histograms of both JSs and average surface distances (ASDs) for P-HNN+RW, with and without the ground-truth mask, and for the PTK. The last two rows of Tbl. 1 provide lobe-specific results for comparison with PTK. As can be seen, when using the same mask, our method (0.89 JS) convincingly outperforms the PTK's (0.84 JS), producing statistically significant improvements ($p < 0.001$ Wilcoxon signed-rank test) for both metrics, despite excluding volumes that the PTK could not segment. Fig. 3 provides some qualitative examples, demonstrating the impact of the P-HNN+RW's improvements in terms of visual quality and segmentation usability.

Table 1: Mean JSs and their standard deviation for each lobe on the our test set. Starred results only use the 114 volumes the PTK could segment, otherwise results include the entire 154 volume test set, which may include more difficult cases. Overall scores average each lobe score, weighted by their relative number of pixels.

| Lobe | LU | LL | RU | RM | RL | Overall |
|---|---|---|---|---|---|---|
| **P-HNN+RW (GT Mask)** | $0.925 \pm 0.175$ | $0.945 \pm 0.116$ | $0.929 \pm 0.057$ | $0.824 \pm 0.142$ | $0.887 \pm 0.239$ | $0.888 \pm 0.164$ |
| **P-HNN+RW*** | $0.930 \pm 0.144$ | $0.925 \pm 0.130$ | $0.927 \pm 0.044$ | $0.848 \pm 0.091$ | $0.873 \pm 0.244$ | $0.888 \pm 0.146$ |
| **PTK*** | $0.884 \pm 0.171$ | $0.853 \pm 0.202$ | $0.835 \pm 0.148$ | $0.683 \pm 0.222$ | $0.865 \pm 0.170$ | $0.838 \pm 0.140$ |

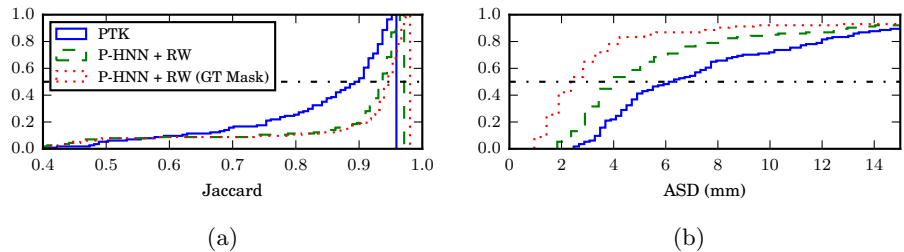

(a)  (b)

Fig. 4: Cumulative histograms of our method's performance on 114 CT scans, *i.e.*, excluding those that the PTK could not segment. (a) and (b) depict JS and ASD results, respectively.

## 4   Conclusion

This work introduces a simple, yet highly powerful, approach to PPLS that uses deep FCNs combined with a RW refinement. We achieve high accuracy without reliance on prior airway or vessel segmentations. Furthermore, the proposed method is developed and validated specifically for pathological lungs. When tested on 154 held-out pathological computed tomography (CT) scans, the largest evaluation of PPLS to date, the proposed method produces strong lobe-specific and overall JSs. Our proposed method also consistently outperforms a state-of-art airway and vessel based PPLS method [6] in both ASD and JS ($p < 0.001$). Importantly, these results are obtained with less complexity relative to the state of the art, increasing our method's generalizability. Thus, our proposed method represents an important step forward toward clinically useful PPLS tools.